\PassOptionsToPackage{table}{xcolor}    
\documentclass[journal,twoside,web]{ieeecolor}
\usepackage{tmi}
\usepackage{cite}
\usepackage{amsmath,amssymb,amsfonts}
\usepackage{algorithmic}
\usepackage{graphicx}
\usepackage{textcomp}

\usepackage{booktabs}
\usepackage{colortbl}

\def\BibTeX{{\rm B\kern-.05em{\sc i\kern-.025em b}\kern-.08em
    T\kern-.1667em\lower.7ex\hbox{E}\kern-.125emX}}
\begin{document}
\title{Robust Activation Map Rectification for \\ Weakly Supervised Volumetric Segmentation: Temporal Coherence as a Free Lunch}
\author{Renshu Gu, Jialiang Chen, Fei Gao, Hang Su, Jun Qi, Jiamin Xu, Yicheng Shen, Jiayu Zhang, \\ Jiaxi Pan, Caiming Zhang, and Gang Xu 
\thanks{Submitted on. This research is supported in part by the National Natural Science Foundation of China (No. U22A2033), and the Open Project of Smart Operations and Control Engineering Technology Center at Civil Aviation Company. \textit{Corresponding author: Gang Xu.} }
\thanks{Renshu Gu, Jialiang Chen, Jiamin Xu, Yicheng Shen, and Gang Xu are with the School of Computer Science, Hangzhou Dianzi University, Hangzhou 310018, China. (email: \{renshugu; 232050115; 251050027; 42746; gxu\}@hdu.edu.cn ) }
\thanks{Fei Gao is with the Hangzhou Institute of Technology, Xidian University, Hangzhou 311231, China. (email: fgao@xidian.edu.cn)}
\thanks{Hang Su is with University of Yamanashi. (email: grainysu11@gmail.com)}
\thanks{Jun Qi is with the Department of Electronic Engineering, School of Information Science and Engineering, Fudan University, Shanghai 200438, China. (email: jqi41@gatech.edu) }
\thanks{Jiayu Zhang and Jiaxi Pan are with the Ruian People's Hospital, Rui'an 325200, China. (email: zhangjiayu0201@wmu.edu.cn; pjx19821213@163.com) }
\thanks{Caiming Zhang is with School of Software, Shandong University, Jinan 250101, China. (email: czhang@sdu.edu.cn) }
}

\maketitle

\begin{abstract}
Weakly supervised segmentation relies heavily on class activation maps (CAMs) to initially localize target regions. However, CAMs are often noisy and prone to catastrophic failures. Existing remedies typically introduce additional training stages or prototype learning, increasing computational cost and reducing robustness. In this paper, we propose a training-free prototype-free framework that rectifies unreliable CAMs by exploiting temporal and structural coherence in volumetric data as a free lunch. Our approach is built on two key components. First, we introduce Variance-Reduced Activation Aggregation (VRAA) which suppresses noise and amplify coherent semantic signals. We provide a theoretical justification by modeling CAMs as high-dimensional random vectors and show that aggregation yields provable variance reduction. Second, we design a Bidirectional Extremity Rectification (BER) mechanism that detects and rectifies implausible activations through bidirectional extremity checks, effectively mitigating extreme-value failures without learning additional parameters. Our method is model-agnostic and can be seamlessly integrated with existing  pipelines. Extensive experiments on multiple public benchmarks demonstrate substantial improvements over state-of-the-art weakly supervised methods, achieving up to $20\%$ Dice and $40\%$ mIoU gains while reducing inference time by more than 5×. These results indicate that leveraging coherence as an implicit inductive bias yields a principled and efficient approach to stabilizing weakly supervised volumetric segmentation. Our code will be available.
\end{abstract}

\begin{IEEEkeywords}
 large foundation model, medical image segmentation, SAM, volumetric medical image, weakly supervised.
\end{IEEEkeywords}

\section{Introduction}
\label{sec:intro}

Volumetric image segmentation \cite{xing2025segmamba,zhang2024cqformer,xu2025gm,11150469} plays a critical role in clinical diagnosis, treatment planning, and disease monitoring. Fully supervised segmentation methods rely on dense pixel-level annotations, which severely limit scalability and practical deployment. As a result, weakly supervised segmentation (WSS), which requires only image-level labels, has emerged as an attractive alternative for clinical use.

A dominant approach in weakly supervised segmentation relies on class activation maps (CAMs) to generate initial localization cues. However, CAMs are inherently noisy and incomplete, and may falsely activate the background. Therefore, much effort has been made to improve the CAMs ~\cite{kweon2021unlocking,lee2021anti, lee2021railroad, xu2021leveraging, du2022weakly,zhou2022regional,hu2023conditional}.

Recently, large foundation models such as the Segment Anything Model (SAM)~\cite{kirillov2023segment} and its medical adaptation MedSAM~\cite{MedSAM} have demonstrated impressive zero-shot segmentation capabilities~\cite{wei2024semantic,li2024asam,xu2024eviprompt,zhang2025alps}. Their strong generalization ability has motivated a new line of research that integrates foundation models into weakly supervised segmentation pipelines. Existing approaches typically fall into two categories. First, many methods seek to transfer knowledge from foundation models via additional training \cite{Lin_2023_CVPR,xie2022clims,kweon2024sam,yang2024foundation}, using CAMs, prototypes, or prompts as intermediate supervision signals. Second, some recent methods explore training-free segmentation \cite{xu2024eviprompt, tang2025towards}, where carefully designed prompts are used to adapt foundation models to unseen tasks.

\begin{figure*}[t]
\centering
\includegraphics[width=1\linewidth]{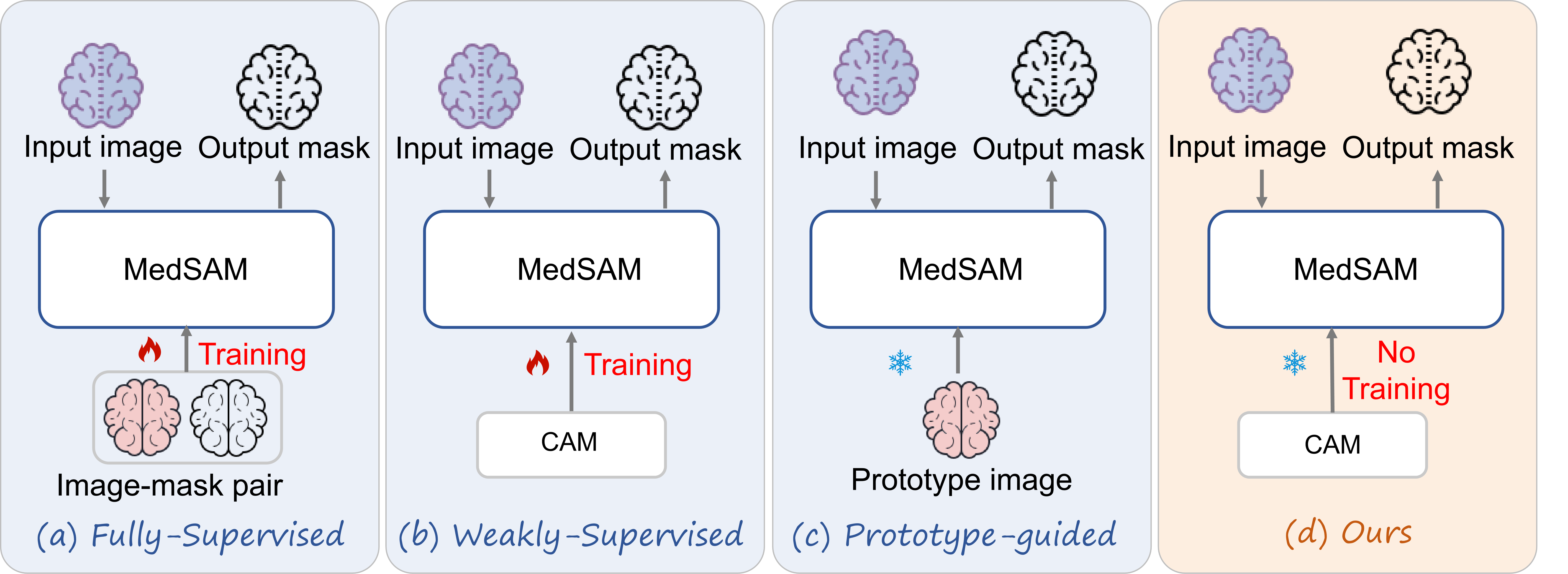}
\caption{\textbf{Different segmentation paradigms using large foundation models (LFMs).} (a) Fully supervised methods. (b) Weakly supervised segmentation frameworks that transfer the knowledge of LFM to the classifier by training, such as S2C ~\cite{kweon2024sam}, FMA-WSSS ~\cite{yang2024foundation}, and CG-CDM ~\cite{10.1007/978-3-031-43901-8_72}. (c) Open-world segmentation, such as IPSeg ~\cite{tang2025towards}, with a prototype image prompt. (d) Our prototype-free weakly supervised segmentation requires no training once the initial CAM is acquired, and is proven to improve both accuracy and efficiency. Temporal coherence is exploited as a free lunch. }
\label{fig:logics} 
\end{figure*}

Despite their promise, these approaches face two fundamental limitations when applied to volumetric medical images. First, training-based methods remain vulnerable to error accumulation: inaccurate CAMs or unstable foundation-model outputs can corrupt the learning process, leading to cascading failures. Second, training-free methods often rely on accurate CAMs or the availability of reliable prototype images. In practice, foundation models may produce unexpected or even implausible segmentation results (catastrophic failures) that contradict basic anatomical or visual common sense.

In this work, we argue that the core challenge of weakly supervised volumetric segmentation in the era of large foundation models is not merely learning better CAMs, but rather stabilizing noisy high-dimensional activation signals before they are consumed by a powerful foundation model. From an image-processing perspective, CAMs can be viewed as noisy observations of an underlying latent structure, in which random fluctuations and outliers often dominate the strongest responses. Naively feeding such unstable cues into a foundation model may lead to suboptimal results.

Motivated by this observation, we propose CSSeg (Common Sense Segmentation), a prototype-free framework for weakly supervised volumetric medical image segmentation that requires no training once initial CAMs are acquired. Instead of introducing additional learning stages, CSSeg focuses on reducing uncertainty and enhancing robustness at the prompt-construction level, and exploits temporal coherence as a free lunch. The framework consists of two key components.

First, we introduce a Variance-Reduced Activation Aggregation (VRAA) strategy that exploits inter-slice coherence to reduce noise variance and thus uncertainty in volumetric data. MedSAM-V2 \cite{zhu2024medical} seeks to train a memory bank to impose temporal consistency. In this regard, we propose a training-free strategy to reduce uncertainty in activation maps. By aggregating CAMs across slices, VRAA suppresses spurious activations and reinforces stable regions that persist throughout the volume. From a high-dimensional perspective, this aggregation reduces noise variance and improves the concentration of meaningful signals, yielding more reliable localization cues.

Second, to address severe failure cases that contradict visual common sense, we propose a Bidirectional Extremity Rectification (BER) mechanism. BER performs extremity-based checks between slice-wise and aggregated activation maps, detecting and correcting frames whose activation patterns violate basic spatial or contextual consistency. This mechanism prevents catastrophic errors from propagating to the foundation model.

CSSeg achieves stable and reliable segmentation without requiring any pixel-level supervision \ref{fig:logics}(a), prototype images\ref{fig:logics}(c), or additional training\ref{fig:logics}(b).It outperforms state-of-the-art weakly supervised and training-free methods by up to 20.5\% Dice and 40.3\% mIoU, while reducing inference time by more than 5×, making it particularly suitable for small or heterogeneous medical datasets.

The contributions of our method can be summarized below.
\begin{itemize}
\item We propose a weakly supervised volumetric image segmentation framework, dubbed CSSeg, that leverages recent large foundation models. The proposed framework requires minimal training and minimal (just one) hyperparameter. 
\item To mitigate unexpected failures produced by large foundation models, we propose a Bidirectional Extremity Rectification (BER) mechanism based on extremity check to prevent the segmentation model from producing unreasonable outputs.
\item To reduce the uncertainty of initial seeds in weakly supervised image segmentation, we impose a Variance-Reduced Activation Aggregation (VRAA) strategy that ensures cross-frame consistency and thus improves reliability.
\item Experimental results on multiple benchmarks show that the proposed framework consistently achieves favorable performance, especially on challenging datasets with limited samples and varying appearance, while offering a significant gain in computational efficiency.
\end{itemize}

\section{Related Work}

\paragraph{Weakly Supervised Segmentation.}
For weakly Supervised Segmentation (WSS), mainstream efforts focus on improving CAMs by adversarial erasing \cite{kweon2021unlocking,lee2021anti}, saliency guidance \cite{lee2021railroad}, affinity learning \cite{ahn2018learning,fan2020cian,xu2021leveraging}, contrast learning \cite{du2022weakly,zhou2022regional}, or conditional diffusion models \cite{hu2023conditional}.

\paragraph{Segmentation with Large Foundation Models.}
With the recent advent of large foundation models, Yang et al. \cite{yang2024foundation} exploit both CLIP and SAM for WSS by prompt learning. They construct an image classification task and a seed segmentation task and learn task-specific prompts to generate high-quality segmentation seeds. Kweon et al. \cite{kweon2024sam} introduce a framework that directly transfers SAM knowledge to the classifier during training via prototype-based contrastive learning and CAM-based prompting. All of the above methods require training to improve the CAMs. A couple of recent methods explore the power of foundation models without training. Xu et al. \cite{xu2024eviprompt} focused on prompt-based adaptation of the SAM model to medical images, whereas we focus on filtering out unreasonable segmentations, and our method is aimed at volumetric segmentation. Recently, Tang et al. \cite{tang2025towards} proposed a training-free framework for open-world segmentation. It requires a prototype (prompt image) of the target object.

\section{Method}
\label{sec:method}

\begin{figure*}[t]
\centering
\includegraphics[width=1\linewidth]{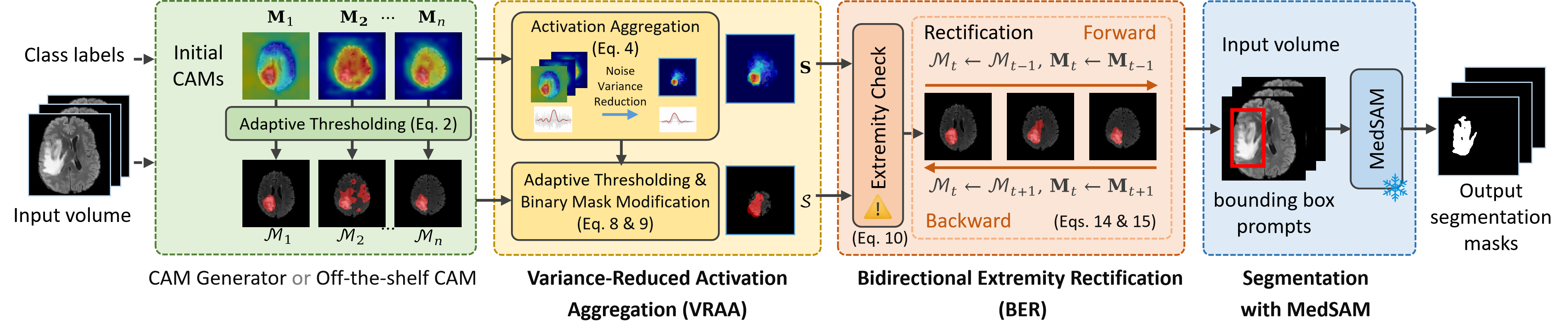}
\caption{\textbf{Overview of the proposed method.} (A) Our overall workflow. Our framework includes the Variance-Reduced Activation Aggregation (VRAA strategy, the Bidirectional Extremity Rectification (BER) module, and postprocessing (Sec. \ref{sec:MedSAM}). (B) Details of the VRAA strategy (Sec. \ref{sec:VRAA}). (C) The BER module. Further details can be found in Sec. \ref{sec:BER}. (D) The MedSAM-based segmentation with postprocessing (Sec. \ref{sec:MedSAM}).}
\label{fig:overall}
\end{figure*}











The overall framework of the proposed method is shown in Fig. \ref{fig:overall}. In Section \ref{sec:CAM}, the initial CAMs are first acquired. Then, Variance-Reduced Activation Aggregation (VRAA) is performed to exploit cross-frame consistency and reduce single-frame uncertainty (Section \ref{sec:VRAA}). Afterwards, Section \ref{sec:BER} introduces the Bidirectional Extremity Rectification (BER) mechanism. Finally, post-processing is performed in Section \ref{sec:MedSAM}.

\subsection{Preliminaries: A Random Matrix Perspective}
\label{sec:CAM}


The input to our method is a 3D image and its corresponding class activation maps (CAMs). Our method can either \textit{use off-the-shelf CAMs} or \textit{generate the CAMs by fine-tuning on the specific dataset with class-level labels}. 

Consider a sample \( \mathbf{X}(0) \) from the distribution \( D(\mathbf{X}|y) \) and the image-level label \( y = \{0, 1\} \) for binary classification. The volumetric (3D) image data is sliced along a specific axis to generate 2D slices \( \mathbf{X}_t \), where $t$ is the frame number. Each slice is classified using a ViT network \cite{dosovitskiy2020vit}  and the CAM is generated using Grad-CAM \cite{selvaraju2017grad}, denoted as \( \mathbf{M}^c \in \mathbb{R}^{H \times W} \). The process involves:

The CAM can be computed as the weighted sum of feature embeddings: 
\begin{equation}
\label{eq:eq3}
\mathbf{M}^c(x, y) = \text{ReLU}\left(\sum_{i=1}^{N} \alpha_i^c \mathbf{F}_i(x, y)\right),
\end{equation}
where \(\mathbf{F} \in \mathbb{R}^{N \times d}\) is the feature embedding from the last layer of ViT, where \( N \) is the number of patches and \( d \) is the embedding dimension. weights \( \alpha_i^c \) are computed from the gradient of the output \( y^c \) for class \( c \). we adopt a Class-aware Attention-based Affinity (CAA) module \cite{Lin_2023_CVPR} to improve the previously generated CAMs. The improved CAM is denoted as \( \mathbf{M}_t \in \mathbb{R}^{H \times W}\).







Typically in the CAM-based method, a threshold \( \tau \in [0,1] \) is applied on the CAM to obtain a segmentation mask \( \mathbf{\mathcal{M}_t} \in \{0,1\}^{H \times W} \) as:
\begin{equation}
\label{eq:init_mask}
{\mathcal{M}_t}(x, y) = 
\begin{cases} 
      1 &  \mathbf{M}_t(x, y) \geq \tau *\max(\mathbf{M}_{t}(x, y)) \\
      0 &  \mathbf{M}_t(x, y) < \tau *\max(\mathbf{M}_{t}(x, y)).
   \end{cases}
\end{equation}
After this procedure, our method requires no further training. \textit{Note that $\tau$ is the only hyperparameter in our entire method.}

In the weakly supervised setting, CAMs are known to be noisy, spatially diffuse, and unstable, particularly in medical images where foreground regions are subtle and heterogeneous. From a Random Matrix Theory (RMT) perspective, a CAM can be viewed as a noisy observation of an underlying latent activation pattern. We formalize a statistical interpretation of slice-wise CAMs that motivates our design. Each CAM is defined as:
\begin{equation}
\label{eq:RMT}
\mathbf{M}_t = \mathbf{M}^{*}_t + \sigma\boldsymbol{\xi}_t,\quad
\mathbf{M}_t,\mathbf{M}^{*}_t\in\mathbb{R}^{d},\; d=H{\times}W.
\end{equation}
where \(\mathbf{M}^{*}_t\) is a latent signal corresponding to the true values,  \(\boldsymbol{\xi}_t\) is a zero-mean isotropic random vector capturing high-dimensional noise, and \( \sigma\) controls noise magnitude.

Spurious CAM activations often arise from the extreme values of \(\boldsymbol{\xi}_t\) rather than meaningful structure, especially when \(d \) is large. Our goal is therefore not to relearn CAMs, but to reduce noise variance and suppress extreme outliers before using them as prompts for a foundation model.

\textbf{Discussion.} We adopt a random matrix perspective because CAMs are high-dimensional objects whose noise components arise from the aggregation of many weakly dependent sources, including feature entanglement and model mismatch. Random matrix theory provides a principled framework for characterizing extreme-value behavior via concentration phenomena, without relying on exact distributional assumptions. More importantly, our analysis only depends on mild conditions such as zero-mean noise. This perspective explains why spurious extrema dominate weakly supervised CAMs in high dimensions and why aggregating coherent activations leads to provable variance reduction.

\subsection{Variance-Reduced Activation Aggregation}
\label{sec:VRAA}

In our case, the uncertainty is quantified by the variance of noise-induced activation fluctuations. Therefore, we propose a simple and effective Variance-Reduced Activation Aggregation (VRAA) method that incorporates cross-slice coherence and reduces uncertainty in a single frame. 



\paragraph{Noise Variance Reduction.} 
Given the CAM $\mathbf{M}_t(t=1,...,n)$ of multiple frames, we perform a normalized aggregation along the $t$ axis. $n$ represents the number of slices from the same 3D volumetric image data: 
\begin{equation}
\mathbf{S}(x, y) = \frac{1}{n} \sum_{t=1}^n {\mathbf{M}_t}(x, y) = \mathbf{S}^*+\frac{\sigma}{n} \sum_{t=1}^{n} \boldsymbol{\xi}_t,
\label{eq:norm_stack}
\end{equation}


\noindent where \(\mathbf{S}\) is the observed aggregated CAM, \(\mathbf{S}^*\) is the ideal latent signal. For simplicity, assume noise \(\boldsymbol{\xi}_t\) in Eq. (\ref{eq:RMT}) are i.i.d. isotropic sub-Gaussian with \(\mathbb{E}(\boldsymbol{\xi}_t) = 0,~ \mathbb{E}(\boldsymbol{\xi}_t \boldsymbol{\xi}_t^\top) = I.\) (For non-i.i.d. noises, similar bounds hold under weak dependence via covariance control, as presented in the Appendix.) Noise vectors concentrate around \(\|\boldsymbol{\xi}_t\|_2 \approx \sqrt{d}\)~\cite{vershynin2018high}, and after averaging, the variance is:
\begin{equation}
\label{eq:var_iid}
\text{Var} \left( \frac{1}{n} \sum_{t=1}^{n} \boldsymbol{\xi}_t \right) = \frac{1}{n^2} \sum_{t=1}^{n} \text{Var}(\boldsymbol{\xi}_t)=\frac{1}{n^2} \cdot nI = \frac{1}{n}I.
\end{equation}


The standard deviation of the aggregated noise scales as \(\sigma/\sqrt{n}\), shrinking by a factor of \(1/\sqrt{n}\), leading to a reduction in uncertainty. Thus:
\begin{equation}
\left\| \frac{1}{n} \sum_{t=1}^{n} \boldsymbol{\xi}_t \right\|_2 \approx \sqrt{\frac{d}{n}},
\end{equation}

\begin{equation}
\|\mathbf{S} - \mathbf{S^*}\|_2 =O\left(\sigma\sqrt{\frac{d}{n}}\right) \quad \text{with high probability}.
\end{equation}

\paragraph{Integrated Binary Mask.} 
Subsequently, for the aggregated heatmap $\mathbf{S}$, to obtain the binary mask $\mathbf{\mathcal{S}} \in \{0,1\}^{H \times W}$ from $\mathbf{S}\in \mathbb{R}^{H \times W}$, the exact same $\tau$ as in Eq. (\ref{eq:init_mask}) is used for adaptive thresholding: 

\begin{equation}
\mathbf{\mathcal{S}}(x, y) = 
\begin{cases} 
      1 &  {\mathbf{S}}(x, y) \geq (1-\tau) *\max({\mathbf{S}}(x, y)) \\
      0 &  {\mathbf{S}}(x, y) < (1-\tau) *\max({\mathbf{S}}(x, y)),
   \end{cases}
\end{equation}

\noindent where $\mathcal{S} \in \mathbb{R}^{ H \times W}$ is the aggregated region of interest. Overall, VRAA enables continuous activation of the target region across the entire 3D space, helping to reduce the impact of random noise in individual slices, and reinforcing consistent signals across slices to highlight any potential target regions. 

\paragraph{Binary Mask Modification.}
$\mathcal{S}$ tends to cover a larger area than the segmentation mask of any individual frame. We calculate the intersection of individual-frame's segmentation masks $\mathcal{M}_t$ with $\mathcal{S}(x, y)$, obtaining a new binary mask for each frame $t$:
\begin{equation}
 \mathcal{R}(\mathcal{M}_t)  = \mathcal{R}({\mathcal{M}}_t) \cap \mathcal{R}({\mathcal{S}}),
\end{equation}
where the region operation is defined as $\mathcal{R}({\mathcal{M}}) = \{(x, y) \mid {\mathcal{M}}(x, y) = 1\}$.

\subsection{Bidirectional Extremity Rectification}
\label{sec:BER}

\subsubsection{Bidirectional Extremity Check}

In some cases, the generated activation map can exhibit unexpected errors, with instances in which the activation region \textit{entirely misses the target region}. In other words, individual CAMs can be dominated by noise-induced extremes. This becomes a major problem in weakly supervised image segmentation enabled by large foundation models.

To address this, we present a Bidirectional Extremity Rectification (BER) module that performs bidirectional extremity checks and rectifies problematic frames using adjacent frames. 
Specifically, \textit{if the following extreme conditions are met, the class activation map for the frame is considered to perform poorly; otherwise, it is regarded as performing fine.} Violations of the extremity check indicate statistical inconsistency between local and global activation supports.

\begin{figure}[t!]
\centering
\includegraphics[width=1\linewidth]{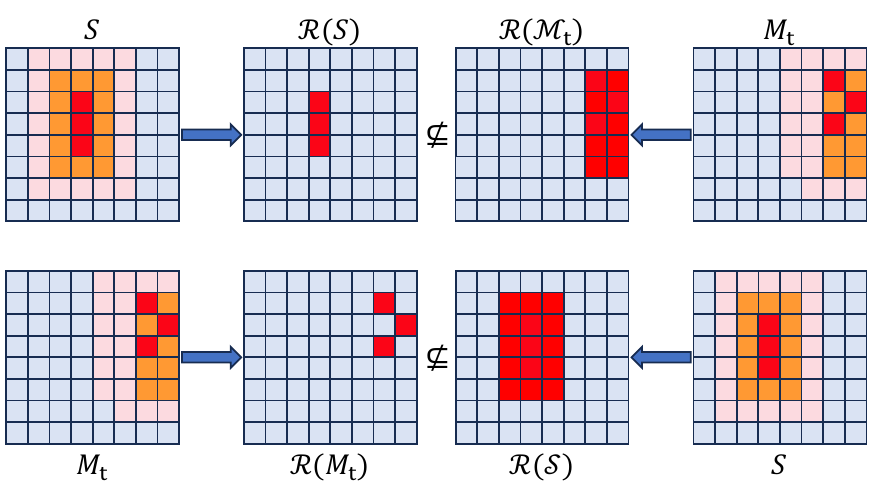}
\caption{\textbf{Bidirectional Extremity check.} The top row checks if the highest responses of the aggregated activation map fall out of the region of frame $t$'s binary mask $\mathcal{M}_t$. The bottom row checks if the highest responses of CAM at frame $t$ fall out of the region of the binary mask $\mathcal{S}$ of the aggregated activation map.}
\label{figure:twoway}
\end{figure}

The extreme conditions are as follows:
\begin{equation}
\label{eq:extreme_cond}
\textcolor{red}{\mathcal{R}({\mathbf{M}}_t) \nsubseteq \mathcal{R}(\mathcal{S}) }\vee
\textcolor{blue}{\mathcal{R}({\mathbf{S}}) \nsubseteq \mathcal{R}(\mathcal{M}_t)}, 
\end{equation}
where, $\vee$ denotes logic OR, \textcolor{red}{$\mathcal{R}({\mathbf{M}}_t) \nsubseteq \mathcal{R}(\mathcal{S})$} means the highest responses of CAM at frame $t$ does not even fall within the region of binary mask $\mathcal{S}$, and \textcolor{blue}{$\mathcal{R}({\mathbf{S}}) \nsubseteq \mathcal{R}(\mathcal{M}_t)$} means the highest response of aggregated activation map does not fall within the region of frame $t$'s binary mask $\mathcal{M}_t$ (shown in Fig. \ref{figure:twoway}). If \textit{any} of the two conditions is met, it indicates that there may be some severe errors.

\subsubsection{Theoretical Basis for the Extremity Check} 
Assume a noise-only CAM: \( \mathbf{M}_{noise} = \sigma \xi, \quad \xi_i \sim \mathcal{N}(0, 1) \). According to the standard random matrix theory \cite{vershynin2018high}, the extreme-value is:
\begin{equation}
\label{eq:max_noise}
\max_{1 \leq i \leq d} \xi_i \approx \sqrt{2 \log d} \quad \text{with high probability},
\end{equation}

\noindent where \(d\) is the number of pixels. Now assume a signal plus noise CAM as defined in Eq. (\ref{eq:RMT}). Let \(A = \max_i s_i \text{ (peak signal strength)}. i^* = \arg \max_i s_i\) is the peak signal location. For the signal peak to dominate the noise maxima, we need: \(A + \sigma \xi_{i^*} > \max_{j \neq i^*} \sigma \xi_j \). Given Eq.(\ref{eq:max_noise}) and \(\xi_{i^*} = O(1) \), then with high probability: 
\begin{equation}
A \gtrsim C \sigma \sqrt{\log d}.
\end{equation}

\noindent After VRAA which reduces noise scale by \(1/\sqrt{n} \), the condition becomes much easier to satisfy:
\begin{equation}
\label{eq:mild_assumption}
A \gtrsim C \sigma \sqrt{\frac{\log d}{n}}.
\end{equation}

\noindent The extremity check (Eq. (\ref{eq:extreme_cond})) works because it leverages the fact that in high-dimensional activation maps, noise-induced extrema scale as \( O(\sigma \sqrt{\log d}) \) and are spatially unstable across slices. Under a mild separation assumption (Eq. (\ref{eq:mild_assumption})), signal-dominated maxima align with the aggregated activation support, whereas noise-dominated maxima violate this consistency.

\subsubsection{Bidirectional Rectification}

Once unreliable frames are detected, the preceding or subsequent frame may offer better segmentation. Consequently, we apply bidirectional optimization, where the activation map $\mathbf{M}_t$ and segmentation mask $\mathcal{M}_t$ are conditionally replaced with the $t-1$ frame (forward process) or $t+1$ frame (backward process). The forward process is defined as:
\begin{small}
\begin{equation}
\label{eq:replace_forward}
\begin{cases} 
\mathcal{M}_t \leftarrow \mathcal{M}_{t-1}\\
\mathbf{M}_t\leftarrow\mathbf{M}_{t-1}
\end{cases}
\kern-1em \Leftarrow
\begin{cases} 
\mathcal{R}({\mathbf{M}}_t) \nsubseteq \mathcal{R}(\mathcal{S}) \vee
\mathcal{R}({\mathbf{S}}) \nsubseteq \mathcal{R}(\mathcal{M}_t)\\
L(\mathbf{X}_{t-1},\mathcal{M}_{t-1})\!>L(\mathbf{X}_{t},\mathcal{M}_{t})
\end{cases},
\end{equation}
\end{small}
where ${\mathbf{X}_t} \in \mathbb{R}^{ H \times W}$ denote the original input image at Frame $t$. $L(\cdot)$ represents the operation used to extract the average brightness value of the image within the mask, indicating the tissue density in medical image. Eq. (\ref{eq:replace_forward}) means that if the extremity condition is met and the previous frame contains a higher-density tissue region, we replace $\mathbf{M}_t$ and $ \mathcal{M}_t$ with the previous frame, since it tends to better segment the target region. Similarly, the backward process is defined as:
\begin{small}
\begin{equation}
\begin{cases} 
\mathcal{M}_t \leftarrow \mathcal{M}_{t+1}\\
\mathbf{M}_t\leftarrow\mathbf{M}_{t+1}
\end{cases}
\kern-1em \Leftarrow
\begin{cases}
\mathcal{R}({\mathbf{M}}_t) \nsubseteq \mathcal{R}(\mathcal{S}) \vee
\mathcal{R}({\mathbf{S}}) \nsubseteq \mathcal{R}(\mathcal{M}_t)\\
L(\mathbf{X}_{t+1},\mathcal{M}_{t+1})\!>L(\mathbf{X}_{t},\mathcal{M}_{t})
\end{cases}.
\end{equation}
\end{small}

The BER module contributes largely to CSSeg's final performance (Dice increased from 0.5227 to 0.6158 in the ablation studies in Table \ref{tab:ablations}). 









\subsection{Segmentation with MedSAM}
\label{sec:MedSAM}

We then extract the bounding box rather than points or a pixel-level mask to feed into MedSAM. According to \cite{cheng2023sammed2d}, improved bounding boxes tend to produce better segmentation results.
Given the segmentation mask $\mathcal{M}_t(x, y)$, all contours $\mathbf{C}_t = \{c_1, c_2, \dots, c_n\}$ are identified. For each contour $c$, we compute its minimum enclosing bounding box $(x_0, y_0, x_1, y_1)$ and output the bounding box matrix for the current slice, $\mathbf{B}_t \in \mathbb{R}^{ |C_t| \times 4}$. $\mathbf{B}_t$ is then used as the bounding box prompt for MedSAM to obtain the class-specific segmentation mask $\mathcal{P}_t \in \{0,1\}^{H\times W}$ as follows:
\begin{equation}
\label{eq:P_from_SAM}
{\mathcal{P}}_t = \text{MedSAM}(\mathbf{X}_t; \mathbf{B}_t).
\end{equation}

Furthermore, we apply a secondary refinement process that incorporates both VRAA and BER. Equation \ref{eq:norm_stack} through \ref{eq:P_from_SAM} is re-implemented for a second time, resulting in our final output. The repeating process is referred to as DoubleSAM in the experiments. It contributes moderately to the final performance (0.6158 to 0.6325 Dice increase in Table \ref{tab:ablations}).

\section{Experiments}
\label{sec:exp}

\subsection{Dataset and Evaluation Metrics}
\textbf{The BraTS dataset} (Brain Tumor Segmentation Challenge)~\cite{baid2021rsna} is a key benchmark for brain tumor segmentation and includes multi-parametric MRI sequences from multiple institutions. The Adult Glioma Challenge features 5,880 MRI scans from 1,470 patients with diffuse gliomas. We exclusively use the FLAIR sequences, treating all tumor types as a single class. The official training set is divided into training, validation, and testing subsets in an 8:1:1 ratio. For preprocessing, each volumetric scan is sliced into 2D images of 224$\times$224 resolution, following the methodology in \cite{hu2023conditional}. 

\textbf{The CHAOS dataset} (Combined CT-MR Healthy Abdominal Organ Segmentation) was introduced in the ISBI 2019 CHAOS Challenge \cite{kavur2021chaos}. It comprises 20 volumes of T2-SPIR MR abdominal scans. Following CG-CDM \cite{10.1007/978-3-031-43901-8_72}, we focus on segmenting the kidneys. We partition the dataset following the same 8:1:1 train-validation-test ratio and report our performance metrics accordingly. The images are resized to 128 $\times$ 128.

\textbf{The MSD-Brain, Prostate and Cardiac datasets} \cite{antonelli2022medical}, namely Medical Segmentation Decathlon, is a recent Generalizable 3D Semantic Segmentation benchmark that offers diverse segmentation tasks. Its aim is to encourage a model that works out-of-the-box on many tasks, with a tremendous impact on healthcare. We evaluate CSSeg on MSD-Brain, MSD-Prostate, and MSD-Cardiac to validate its strong generalisability.

\textbf{Statement:} Informed consent was obtained in the original studies for all datasets used, and data access and usage in this work comply with the respective dataset agreements/licenses.




\begin{figure*}[t]
\centering
\includegraphics[width=1\linewidth ]{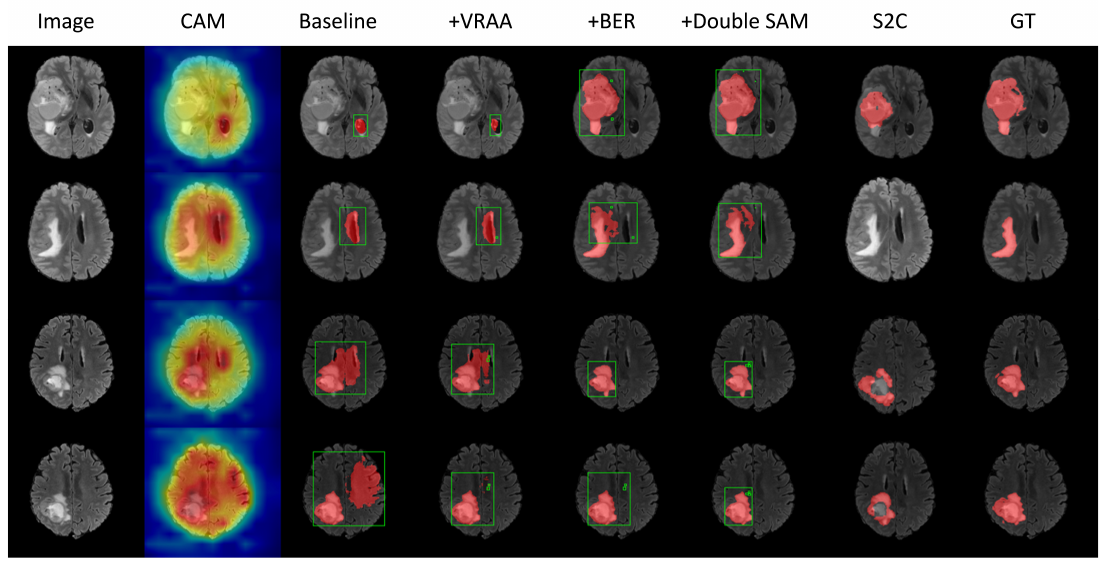}
\caption{\textbf{Segmentation performance on the BraTS dataset.} Red color highlights the segmented area. From left to right: input image; input CAM; Baseline (feeding the initial CAMs to MedSAM.); Baseline + VRAA (Sec. \ref{sec:VRAA}); Baseline + VRAA + BER (Sec. \ref{sec:BER}); Baseline + VRAA+ BER + DoubleSAM (Sec. \ref{sec:MedSAM}); result of S2C\cite{kweon2024sam}; ground truth (GT). The baseline model is prone to mistakenly identifying irrelevant areas as the target, whereas our framework can successfully reduce uncertainty and identify the correct target only.}
\label{figure:eval_BraTS}
\vspace{-0.3cm}
\end{figure*}

\subsection{Implementation Details}
\label{sec:Implementation Details}
For CAM generation, the sole component requiring training in our framework, we retain flexibility by allowing any externally generated CAM results to be used directly as input. Specifically, we use a ViT-B-16 model pre-trained on ImageNet-21k \cite{dosovitskiy2020vit} and extract the output of its last self-attention layer before the classification head. The hyperparameter $\tau$ is set to 0.8 for all benchmarks. Segmentation performance is not sensitive to $\tau$; $\tau$ ranging from 0.5 to 0.8 results in robust performance. We use the official MedSAM-ViT-B model \cite{MedSAM} and conduct all experiments on a single NVIDIA RTX 4090 GPU.

\subsection{ Quantitative and Qualitative Evaluation}
\label{sec:comparison}
\subsubsection{Results on BraTS} 
\textbf{Metrics.} We use the Dice Similarity Coefficient (DSC) to quantify mask overlap, the Mean Intersection over Union (mIoU) to assess region-wise correspondence, and the 95th-percentile Hausdorff Distance (HD95) to assess boundary-level accuracy against ground-truth annotations.

Table \ref{tab:Segmentation Comparison} shows quantitative comparisons on the BraTS dataset. We compare with several weakly-supervised image segmentation approaches that reported results on BraTS, including CG-Diff \cite{Wolleb2022DiffusionMF} and CG-CDM \cite{10.1007/978-3-031-43901-8_72}. Furthermore, we compare against multiple open-source, state-of-the-art segmentation methods, Clip-ES \cite{Lin_2023_CVPR}, IPSeg \cite{tang2025towards}, S2C \cite{kweon2024sam} and SAM 3 \cite{carion2025sam3segmentconcepts} by running their provided code on BraTS. Our method outperforms the second-best method (CG-CDM \cite{10.1007/978-3-031-43901-8_72}) by 12.4\% in Dice and 13.3\% in mIoU. Results show that the state-of-the-art weakly-supervised image segmentation method Clip-ES\cite{Lin_2023_CVPR} does not adapt well to medical images, whereas the proposed method adapts well to medical images. S2C \cite{kweon2024sam} produces suboptimal segmentation on medical images. Furthermore, existing training-free open-set segmentation method IPSeg \cite{tang2025towards} may not have the best performance either. We not only test the official code of S2C and IPSeg \cite{tang2025towards} but also replace SAM in their frameworks with MedSAM, which yields only marginal performance gains. In contrast, our method can produce good segmentation results exploiting MedSAM, while requiring no pixel-level labels.

Figure \ref{figure:eval_BraTS} demonstrates the qualitative evaluation on BraTS. The baseline is to feed the initial CAMs to MedSAM. The 6th column (Baseline + VRAA + BER + DoubleSAM) shows our final output. Through Row 1 to 4, our method outperforms S2C. In Row 2, S2C misses the entire region, while our method segmented the target area. 
\begin{table}[t]
    \centering
    \tabcolsep=6pt
    \caption{\textbf{Comparison of segmentation methods on the BraTS dataset.} Best in bold. The second-best results are marked by underscores. $\uparrow$ means the higher the better, and $\downarrow$ the lower the better.}
    \label{tab:Segmentation Comparison}
    \begin{tabular}{l|l|ccc}
    \toprule
        Method &Publication & \text{Dice} $\uparrow$ & \text{mIoU} $\uparrow$ & \text{HD95} $\downarrow$  \\
        \midrule
        GradCAM & ICCV'17 & 0.235 & 0.149 & 44.4 \\
        GradCAM++ &WACV'18 & 0.281 & 0.187 & 32.6 \\
        ScoreCAM &CVPRW'19 & 0.303 & 0.202 & 32.7 \\
        LayerCAM &TIP'21 & 0.276 & 0.184 & 30.4 \\
        CG-Diff &MICCAI'22 & 0.456 & 0.325 & 43.4 \\
        CG-CDM &MICCAI'23 & \underline{0.563} & \underline{0.450} & \underline{19.2} \\
        Clip-ES & CVPR'23 & 0.148 & 0.083 & 89.6 \\
        S2C  (SAM) & CVPR'24 & 0.028 &0.014 & 74.1 \\
        S2C  (MedSAM) & CVPR'24 & 0.105 & 0.101 &70.1 \\
        IPSeg(SAM) &IJCV'25& 0.170 &0.096 &84.4\\
        IPSeg(MedSAM) &IJCV'25& 0.012 &0.006 &72.7 \\
        SAM 3 & arXiv'25 & 0.482 & 0.377 & 24.3 \\
        \midrule
       \rowcolor[gray]{0.95} 
       CSSeg (Ours) & & \textbf{0.633} & \textbf{0.510} & \textbf{18.7}\\
        \bottomrule
    \end{tabular}
\end{table}





\subsubsection{Results on CHAOS} 
\begin{figure*}[h]
\centering
\includegraphics[width=1\linewidth]{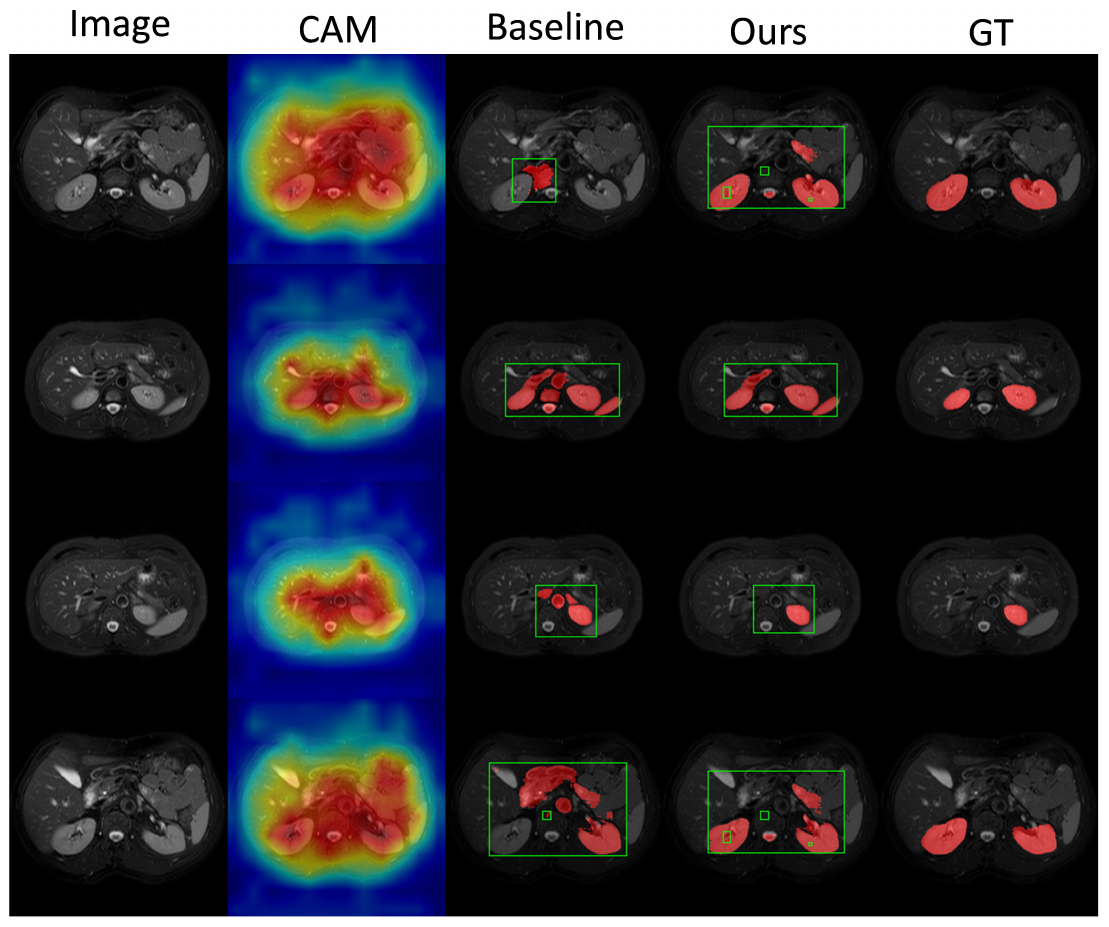}
\caption{\textbf{Qualitative evaluation on CHAOS.} From left to right: input image; initial CAM; baseline; CSSeg result; ground truth. CSSeg corrects noisy initial CAMs, yielding improved segmentation results on this challenging dataset.}
\label{figure:chaosxiaoguo}
\end{figure*}

Table \ref{tab:Segmentation_Comparison_Chaos} presents the quantitative evaluation on CHAOS. Our method is compared against state-of-the-art weakly-supervised approaches, including CG-Diff 
 \cite{Wolleb2022DiffusionMF}, CG-CDM \cite{10.1007/978-3-031-43901-8_72}, and CLIP-ES \cite{Lin_2023_CVPR}. On this challenging dataset, our method outperforms the second-best CG-CDM \cite{10.1007/978-3-031-43901-8_72} by 20.5\% Dice and 40.3\% mIoU. For such a dataset with a limited sample size and diverse appearance across slices, it is essential to obtain reliable prompts with lower uncertainty and to filter out as much noise as possible. Otherwise, large foundation models might be subject to a catastrophic failure. Figure \ref{figure:chaosxiaoguo} shows our segmentation performance on CHAOS. Results indicate that our method can produce fair segmentation masks even when the original CAMs are visually indistinct. Meanwhile, there remains room for improvement on this challenging dataset.

\begin{table}[t]
    \centering
    \tabcolsep=2pt
    \caption{\textbf{Comparison of segmentation methods on the CHAOS dataset.} Best in bold. The second is marked by underscores. $\uparrow$ means the higher the better, and $\downarrow$ the lower the better.}
    \label{tab:Segmentation_Comparison_Chaos}
    \begin{tabular}{l|l|ccc}
    \toprule
        Method & Publication & \text{Dice} $\uparrow$ & \text{mIoU} $\uparrow$ & \text{HD95} $\downarrow$  \\
        \midrule
        GradCAM & ICCV'17 & 0.105 & 0.059 & 33.9 \\
        GradCAM++ & WACV'18 & 0.147 & 0.085 & 28.5 \\
        ScoreCAM & CVPRW'19 & 0.135 & 0.078 & 32.1 \\
        LayerCAM & TIP'21 & 0.194 & 0.131 & 29.7 \\
        CG-Diff & MICCAI'22 & 0.235 & 0.152 & 27.1 \\
        CG-CDM & MICCAI'23 & \underline{0.311} & \underline{0.186} & \underline{23.3} \\
        Clip-ES & CVPR'23 & 0.099 & 0.053 & 88.8 \\
        IPSeg(SAM) & IJCV'25 & 0.096 & 0.051 & 90.0 \\
        IPSeg(MedSAM) & IJCV'25 & 0.004 & 0.002 & 73.2 \\
        \midrule
       \rowcolor[gray]{0.95} 
        CSSeg (Ours) & & \textbf{0.375} & \textbf{0.261} & \textbf{20.2} \\
        \bottomrule
    \end{tabular}
\end{table}

\begin{table}[h]
    \centering
    \caption{\textbf{Comparison of segmentation methods on the MSD-Heart, MSD-Prostate and MSD-Brain datasets.} }
    \label{tab:Segmentation_Comparison_MSD}
    \begin{tabular}{l|l|ccc}
    \toprule
        Method &  Publication & Heart & Prostate & Brain \\
        \midrule
        
        Grad-cam & ICCV'17  &0.059 &0.143 &0.215 \\

        Clip-ES & CVPR'23    &0.030  & 0.029 &0.171\\
        MedSAM-V2& arXiv'25  &0.022 &0.036 &0.133 \\
        SAM-Med2D & arXiv'23   & 0.148& 0.252 &0.306 \\
        \midrule
        \rowcolor[gray]{0.95} 
       CSSeg(Ours) & &\textbf{ 0.242} &\textbf{0.433}&\textbf{0.467}  \\
        \bottomrule
    \end{tabular}
    \label{tab:more_results}
\end{table}

\begin{figure}[h]
\centering
\includegraphics[width=0.9\linewidth]{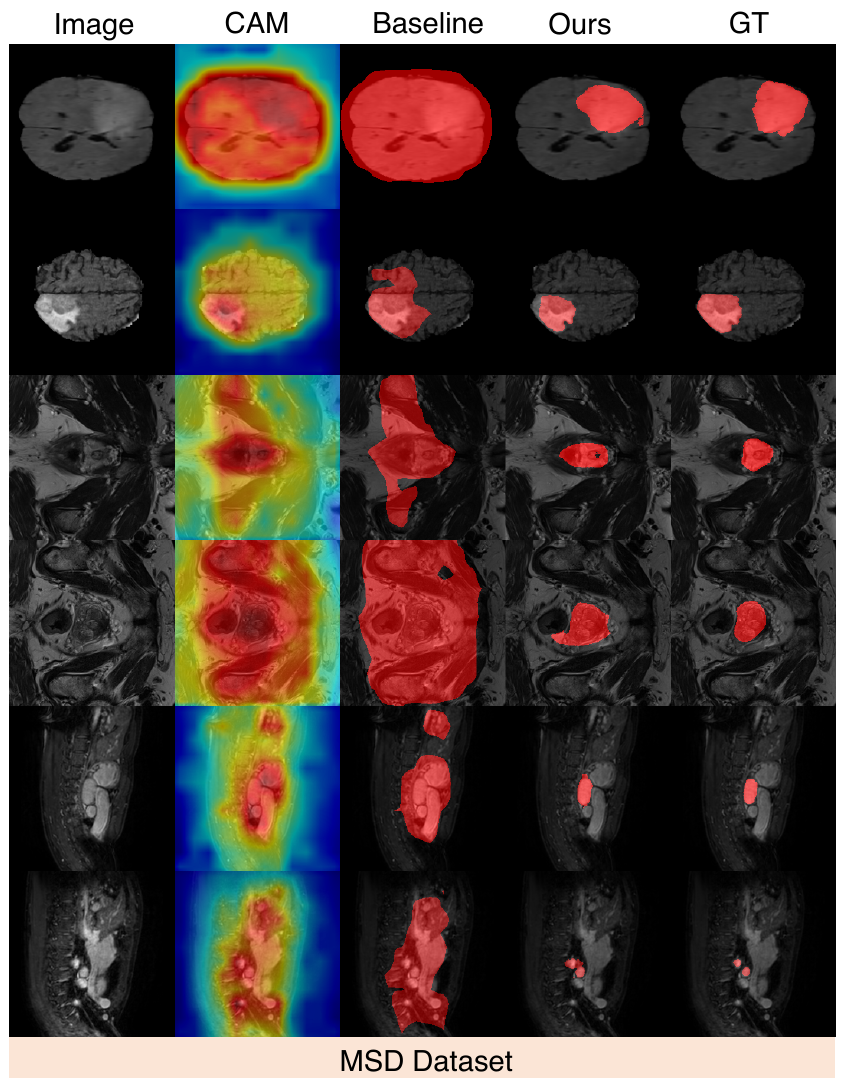}
\caption{\textbf{Qualitative evaluation on MSD.} From let to right: input image; initial CAM; baseline; CSSeg result; ground truth. Our method significantly outperforms the baseline across MSD-Brain (top 2 rows), MSD-Prostate (middle 2 rows), and MSD-Cardiac (bottom 2 rows).}
\label{figure:MSD}
\end{figure}

 \subsubsection{Results on MSD} 
 We further compare our method with state-of-the-art segmentation methods on the Medical Segmentation Decathlon (MSD) \cite{antonelli2022medical} in Table \ref{tab:Segmentation_Comparison_MSD}. We compare with state-of-the-art methods Grad-cam\cite{selvaraju2017grad}, Clip-ES \cite{Lin_2023_CVPR}, MedSAM-V2\cite{MedSAM2} and SAM-Med2D\cite{cheng2023sammed2d}. The proposed CSSeg shows robust generalization ability across diverse tasks, outperforming existing methods. Under minimum training conditions, CSSeg can reach a Dice score of 0.433 when segmenting the prostate central gland and peripheral zone. For the left atrium in cardiac (heart) structures, its Dice score is 0.242, while MedSAM-V2 gives 0.022. For brain tumors, CSSeg achieves a Dice score of 0.467. 
 
 We visualize the segmentation performance and provide qualitative comparisons in Figure \ref{figure:MSD}. The first 2 rows show results for brain tumors, the middle 2 rows show results for prostate zones, and the last 2 rows show segmentation performance for cardiac structures. Compared to the baseline, the segmentation results of our method are substantially more accurate.



\begin{table}[t]
    \centering
    \small 
    \tabcolsep=3.5pt 
    \caption{\textbf{Ablation study of key components on the BraTS dataset.}}
    \label{tab:Ablation Studies}
    \begin{tabular}{c c c c c c}
    \toprule
         VRAA & BER & Double SAM & Dice $\uparrow$ & mIoU $\uparrow$ & HD95 $\downarrow$ \\
    \midrule
         & & & 0.4841 & 0.3821 & 34.19 \\
         \checkmark & & & 0.5227 & 0.4181 & 26.96 \\
         \checkmark & & \checkmark & 0.5223 & 0.4178 & 26.91 \\
         \checkmark & \checkmark & & 0.6158 & 0.4923 & 20.00 \\
         & \checkmark & & 0.6149 & 0.4916 & 20.92 \\
         & \checkmark & \checkmark & 0.6285 & 0.5071 & 20.20 \\
    \rowcolor[gray]{0.9} \checkmark & \checkmark & \checkmark & \textbf{0.6325} & \textbf{0.5104} & \textbf{18.66} \\
    \bottomrule
    \end{tabular}
    \label{tab:ablations}
\end{table}

\subsection{Ablation Studies}
\label{sec:Ablation Studies}




\textbf{Effectiveness of different modules.}  Table \ref{tab:Ablation Studies} elucidates the performance impact of various combinations of components, providing valuable insights into the robustness and efficiency of our model design. Our baseline is feeding the initial CAM into MedSAM. The VRAA module considerably improves segmentation performance (+0.038 Dice), whereas BER yields the largest performance gain (+0.093 Dice). 

\noindent \textbf{Efficiency.} we also compare our inference time with the training-free method IPSeg in Table \ref{tab:infer_time}. Our inference time is 0.393s per image, which is significantly more efficient than IPSeg. Temporal coherence is exploited as a free lunch and adds little computational overhead.

\begin{figure}[t]
\centering
\includegraphics[width=0.8\linewidth]{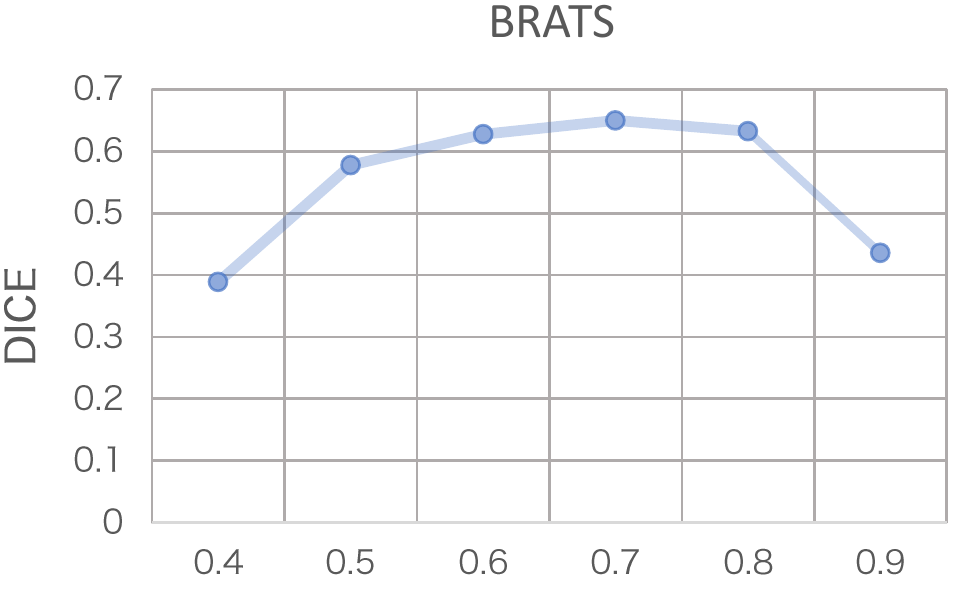}
\caption{\textbf{Ablation study on hyperparameter $\tau$.}}
\label{figure:hyper}
\end{figure}

\noindent \textbf{Sensitivity of $\tau$.} The segmentation performance is not sensitive to $\tau$; a $\tau$ ranging from 0.5 to 0.8 will lead to reasonable segmentation performance, as shown in Fig. \ref{figure:hyper}.

\begin{table}
\caption{\textbf{Ablation study on computational time.} We compare our method with the state-of-the-art training-free segmentation method IPSeg \cite{tang2025towards}. Unit: second per image.}
\begin{center}
\begin{tabular}{lccc} 
\toprule
Method  & Dice$\uparrow$ & Training & Inference \\
\midrule
IPSeg \cite{tang2025towards}  & 0.170  & 0 & 2.144s\\
CSSeg (Ours)  &  \textbf{0.633} & 0 &  \textbf{0.393s}\\ 
\bottomrule
\end{tabular}
\label{tab:infer_time}
\end{center}
\end{table}

\begin{figure}[h!]
\centering
\includegraphics[width=1\linewidth]{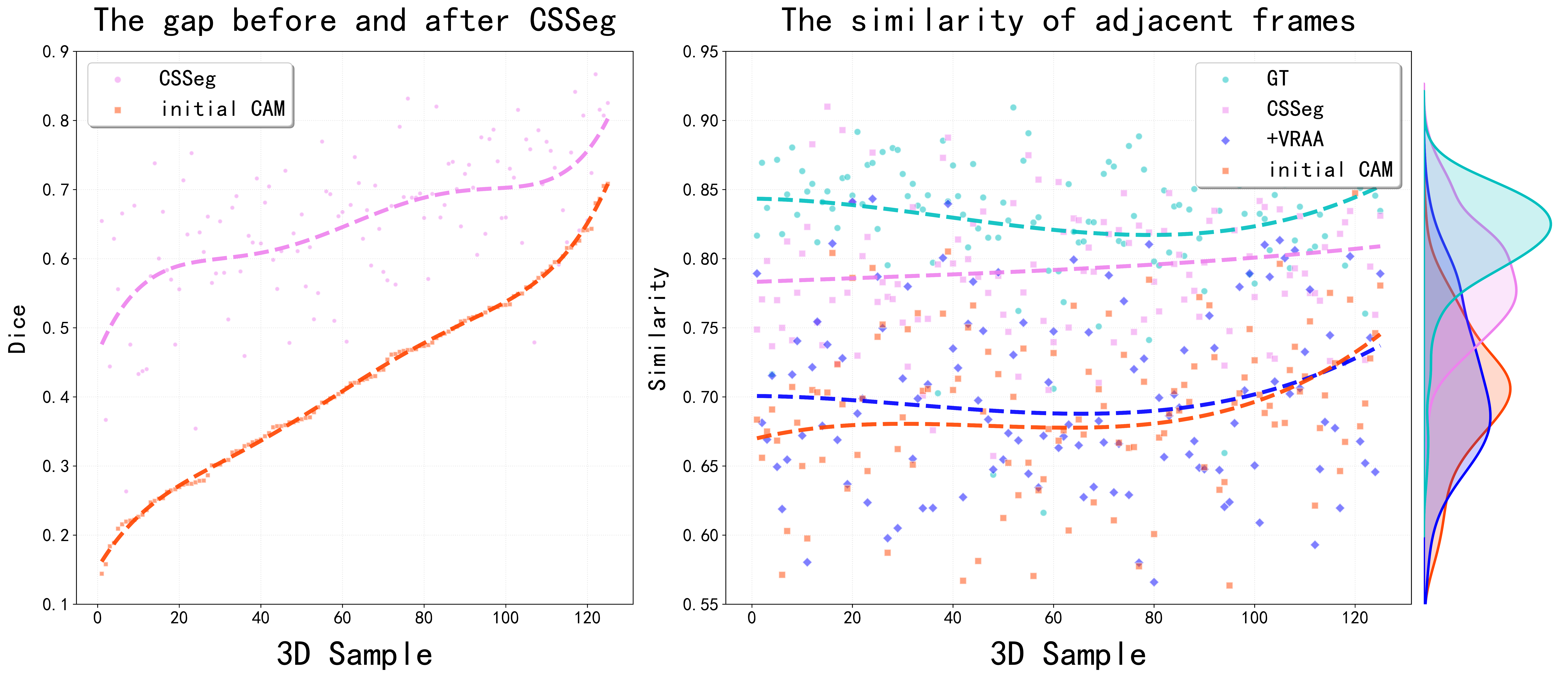}
\caption{\textbf{Robustness to inaccurate initial CAMs.} (a) We sort the 3D samples by initial CAM quality. CSSeg improves Dice over initial CAM consistently. (b) Distribution of adjacent-frame coherence is closer to GT after VRAA and BER. } 
\label{figure:robustness}
\end{figure}

\begin{figure}[h!]
    \centering
    \includegraphics[width=1\linewidth]{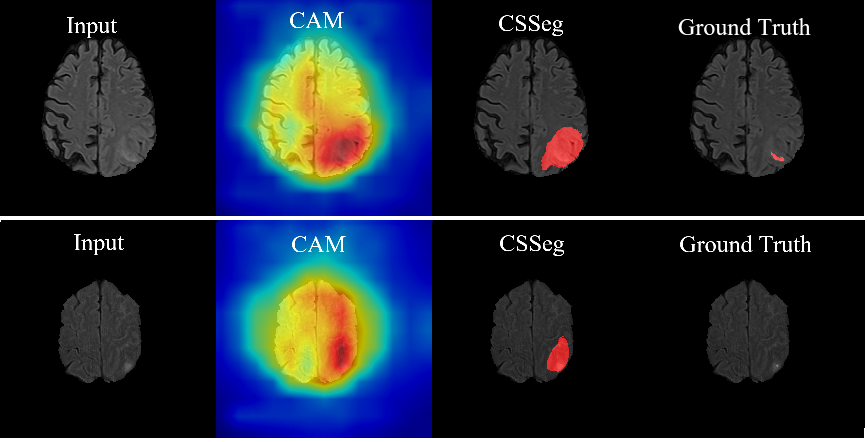}
    \caption{\textbf{Failure cases.} When the target is too small, CSSeg is prone to failure.}
\label{fig:failure_case}
\end{figure}

\subsection{Discussions}
\label{sec:Discussions}
\textbf{Robustness to inaccurate initial CAMs.} We analyze the performance correlated with the accuracy of initial CAMs in Fig. \ref{figure:robustness}(a). We sort the 3D samples by initial CAM quality. The x axis is the samples sorted by the quality of the initial CAMs measured by Dice, and the y axis is the output segmentation results in Dice. Inaccurate CAMs yield downgraded performance, yet even for severely inaccurate CAMs, CSSeg will not collapse but will improve Dice scores to a certain extent. In BraTS, 15 out of 125 3D samples have more than 50\% frames of inaccurate CAMs (Dice < 0.2). These samples have 0.04-0.50 Dice improvement after CSSeg.

Recall that CSSeg makes the following assumptions: multi-frame consistency is desired, and adjacent frames shall not change abruptly. In Fig. \ref{figure:robustness}(b), the similarity of adjacent frames is analyzed. Ground truth (GT) shows high similarity across the frames, and our CSSeg increases adjacent-frame similarity after AMS and BER respectively, which validates our assumption.

\textbf{Limitations.} When the target is too small for CAM to identify, CSSeg is prone to failure (low Dice score) because CAM tends to highlight a larger area, as shown in Fig. \ref{fig:failure_case}. This is due to the intrinsic limitations of CAM, and all CAM-based methods will suffer from this imperfection.
On the other hand, these failure cases may still be clinically meaningful. For tumor diagnosis, for instance, CSSeg highlights a larger area that contains the target.

\section{Conclusion}
\label{sec:conclusions}
This paper proposes a prototype-free weakly supervised volumetric image segmentation framework with minimum training based on large foundation models such as MedSAM. Unlike existing weakly supervised frameworks that train to improve the initial seed CAMs, or existing open-set segmentation paradigms that require a prototype, our method seeks to improve the prompt to MedSAM from a random matrix theory perspective.  Specifically, we design a provable strategy to reduce activation map uncertainty and generate reliable CAMs. Moreover, we propose a Bidirectional Extremity Rectification mechanism to mitigate mistakes that contradicts common sense. Experiments on the BraTS, CHAOS and MSD benchmarks demonstrate that our method performs favorably against state-of-the-art methods without bells and whistles while being computationally efficient, especially for datasets with small sample size and diverse appearance.

\appendices
\section*{Appendix}
\textbf{Beyond i.i.d. Assumption in Eq. \ref{eq:var_iid}.}
In Section \ref{sec:VRAA}, Eq. \ref{eq:var_iid} uses a scalar variance because it implicitly assumes that each activation map is an isotropic i.i.d. random vector. Below are standard ways to keep essentially the same kind of upper bound when $\xi_t$ is not i.i.d. (e.g., temporally correlated across slides). 

Note that variance is sufficient only when noise samples are independent; once samples are correlated (non-i.i.d.), cross-terms no longer vanish, and the correct object controlling concentration is the covariance matrix, not the scalar variance.

{\textit{Let $\xi_t \in \mathbb{R}^d$ be zero-mean but not independent}}. Then, 
\begin{equation}
\operatorname{Cov}\left(\frac{1}{n}\sum\limits_{t=1}^{n}\xi_{t} \right) = \frac{1}{n^2}\sum\limits_{t=1}^{n}\sum\limits_{s=1}^{n} \operatorname{Cov}(\xi_t, \xi_s). 
\end{equation}

So instead of $\frac{1}{n}I$, we can upper bound it by controlling cross-covariances. A clean spectral-norm bound is:
\begin{equation}
\lVert \operatorname{Cov}\left( \frac{1}{n}\sum\limits_{t=1}^{n} \xi_t \right) \rVert_2 \le \frac{1}{n^2} \sum\limits_{t, s} \lVert \operatorname{Cov}(\xi_t, \xi_s) \rVert_2. 
\end{equation}

If we assume a uniform correlation envelope such as:
\begin{equation}
\left\lVert \operatorname{Cov}(\frac{1}{n}\sum\limits_{t=1}^{n} \xi_t) \right\rVert_2 \le \frac{\Sigma_0}{n} \left(1 + 2\sum\limits_{k=1}^{n-1}\left(1 - \frac{k}{n}\right)\rho(k) \right). 
\end{equation}

\textit{Let $\xi$ is stationary with autocovariance matrices $\mathcal{T}_k:=\operatorname{Cov}(\xi_t, \xi_{t+k})$}, then 
\begin{equation}
\operatorname{Cov}\left( \frac{1}{n}\sum\limits_{t=1}^{n}\xi_t  \right) = \frac{1}{n}\mathcal{T}_0 + \frac{2}{n}\sum\limits_{k=1}^{n-1}\left( 1 - \frac{k}{n} \right)\mathcal{T}_k. 
\end{equation}

So, a sufficient assumption is a summability condition
\begin{equation}
\sum\limits_{k\ge 1}\lVert \mathcal{T}_k \rVert_2 < \infty. 
\end{equation}

Then, 
\begin{equation}
\left\lVert \operatorname{Cov}\left( \frac{1}{n} \sum\limits_{t=1}^{n} \xi_t \right) \right\rVert_2 \le \frac{C}{n}, \hspace{2mm} C:=\lVert \mathcal{T}_0 \rVert_2 + 2\sum\limits_{k\ge 1} \lVert \mathcal{T}_k \rVert_2. 
\end{equation}

\bibliographystyle{IEEEtran}

\end{document}